\documentclass[sigconf]{acmart}

\usepackage{float}
\usepackage{graphicx}
\usepackage{amsmath}
\usepackage{booktabs}
\usepackage{algorithm}
\usepackage{algorithmic}
\usepackage{epsfig}
\usepackage{amsmath}
\usepackage{enumitem}
\usepackage{subcaption}
\usepackage{adjustbox}
\usepackage{multirow}
\usepackage{bm}
\usepackage{tabularx}
\usepackage{array}
\usepackage[english]{babel}
\newcommand{\vect}[1]{\boldsymbol{#1}}

\DeclareMathAlphabet\mathbfcal{OMS}{cmsy}{b}{n}

\AtBeginDocument{%
 }

\copyrightyear{2024}
\acmYear{2024}
\setcopyright{rightsretained}
\acmConference[KDD '24]{Proceedings of the 30th ACM SIGKDD Conference on Knowledge Discovery and Data Mining}{August 25--29, 2024}{Barcelona, Spain}
\acmBooktitle{Proceedings of the 30th ACM SIGKDD Conference on Knowledge Discovery and Data Mining (KDD '24), August 25--29, 2024, Barcelona, Spain}
\acmDOI{10.1145/3637528.3672020}
\acmISBN{979-8-4007-0490-1/24/08}

\begin{document}

\title{CAT: Interpretable Concept-based Taylor Additive Models}

\author{Viet Duong$^*$}
\thanks{$^*$Part of this work was done while the author was an intern at AT\&T CDO}
\orcid{0000-0002-0488-8735}
\affiliation{%
\institution{William \& Mary}
\city{Williamsburg}
\state{VA}
\country{United States}}
\email{vqduong@wm.edu}

\author{Qiong Wu}
\orcid{0000-0001-7724-8221}
\affiliation{%
\institution{AT\&T CDO}
\city{Bedminster}
\state{NJ}
\country{United States}}
\email{qw6547@att.com}

\author{Zhengyi Zhou}
\orcid{0009-0004-2866-4746}
\affiliation{%
\institution{AT\&T CDO}
\city{Bedminster}
\state{NJ}
\country{United States}}
\email{zz547k@att.com}

\author{Hongjue Zhao$^\dagger$}
\thanks{$^\dagger$Corresponding author.}
\orcid{0009-0007-8501-6982}
\affiliation{%
\institution{University of Illinois at Urbana-Champaign}
\city{Champaign}
\state{IL}
\country{United States}}
\email{hongjue2@illinois.edu}

\author{Chenxiang Luo}
\orcid{0009-0003-3866-5200}
\affiliation{%
\institution{William \& Mary}
\city{Williamsburg}
\state{VA}
\country{United States}}
\email{cluo02@wm.edu}

\author{Eric Zavesky}
\orcid{0009-0009-5016-2415}
\affiliation{%
\institution{AT\&T CDO}
\city{Austin}
\state{TX}
\country{United States}}
\email{ez2685@att.com}

\author{Huaxiu Yao}
\orcid{0000-0002-8691-9629}
\affiliation{%
\institution{The University of North Carolina at Chapel Hill}
\city{Chapel Hill}
\state{NC}
\country{United States}}
\email{huaxiu@cs.unc.edu}

\author{Huajie Shao$^\dagger$}
\orcid{0000-0001-7627-5615}
\affiliation{%
\institution{William \& Mary}
\city{Williamsburg}
\state{VA}
\country{United States}}
\email{hshao@wm.edu}

\renewcommand{\shortauthors}{Viet Duong et al.}

\begin{abstract}
As an emerging interpretable technique, Generalized Additive Models (GAMs) adopt neural networks to individually learn non-linear functions for each feature, which are then combined through a linear model for final predictions. Although GAMs can explain deep neural networks (DNNs) at the feature level, they require large numbers of model parameters and are prone to overfitting, making them hard to train and scale. Additionally, in real-world datasets with many features, the interpretability of feature-based explanations diminishes for humans. To tackle these issues, recent research has shifted towards concept-based interpretable methods. These approaches try to integrate concept learning as an intermediate step before making predictions, explaining the predictions in terms of human-understandable concepts. However, these methods require domain experts to extensively label concepts with relevant names and their ground-truth values. In response, we propose CAT, a novel interpretable Concept-bAsed Taylor additive model to simplify this process. CAT does not require domain experts to annotate concepts and their ground-truth values. Instead, it only requires users to simply categorize input features into broad groups, which can be easily accomplished through a quick metadata review. Specifically, CAT first embeds each group of input features into one-dimensional high-level concept representation, and then feeds the concept representations into a new white-box Taylor Neural Network (TaylorNet). The TaylorNet aims to learn the non-linear relationship between the inputs and outputs using polynomials. Evaluation results across multiple benchmarks demonstrate that CAT can outperform or compete with the baselines while reducing the need of extensive model parameters. Importantly, it can effectively explain model predictions through high-level concepts. Source code is available at~\href{https://github.com/vduong143/CAT-KDD-2024}{\ttfamily github.com/vduong143/CAT-KDD-2024}.
\end{abstract}





\maketitle
\section{Introduction}
While deep neural networks (DNNs) have demonstrated remarkable success in various areas, the lack of interpretability impedes their deployment in high-stakes applications, such as autonomous vehicles, finance, and healthcare~\cite{arrieta2020explainable}. Thus, enhancing DNN interpretability has emerged as a pivotal area of research in recent years.

Earlier studies primarily focused on perturbation-based post-hoc approaches~\cite{madsen2022post, carvalho2019machine,ivanovs2021perturbation}, but these methods are either computationally expensive or hard to faithfully represent the model's behavior~\cite{ghorbani2019interpretation, rudin2019stop, slack2020fooling}. To address these issues, recent works have shifted focus to Generalized Additive Models (GAMs)~\cite{agarwal2021neural, chang2021node, dubey2022scalable, radenovic2022neural}. GAMs aims to learn non-linear transformation of input features separately into smoothed structures known as shape functions, and then use a linear combination of these functions to make predictions. Despite their potential, GAMs require extensive model parameters and suffer from scalability issue, since they use separate DNNs or an ensemble of numerous decision trees~\cite{lou2012intelligible, chang2021node} to learn the shape function of each feature. Furthermore, while GAMs offer insights into the significance and behavior of individual features via shape function visualizations, these explanations may not always be readily interpretable for humans.

In contrast, human explanations often rely on concept-based reasoning, which semantically groups low-level features into broader concepts, and then explains decisions using these high-level concepts. For example, in medical diagnostics like diabetes, physicians usually explain their conclusions by referring to high-level factors, such as family history, medical history, dietary patterns, and blood tests. This approach has spurred research into integrating concept-based interpretability into DNNs. Specifically, concept-based interpretable methods introduce an intermediate step to learn human-understandable concepts from input features, which then inform predictions made by white-box predictors like linear models or decision trees. Yet, these methods require domain experts to label extensive concepts and their ground-truth values, e.g., categorizing blood test results on a scale from 0 to 71 using the APACHE II scoring system~\cite{apache1985apache}.

\begin{figure*}
    \centering
    \includegraphics[width=.95\textwidth]{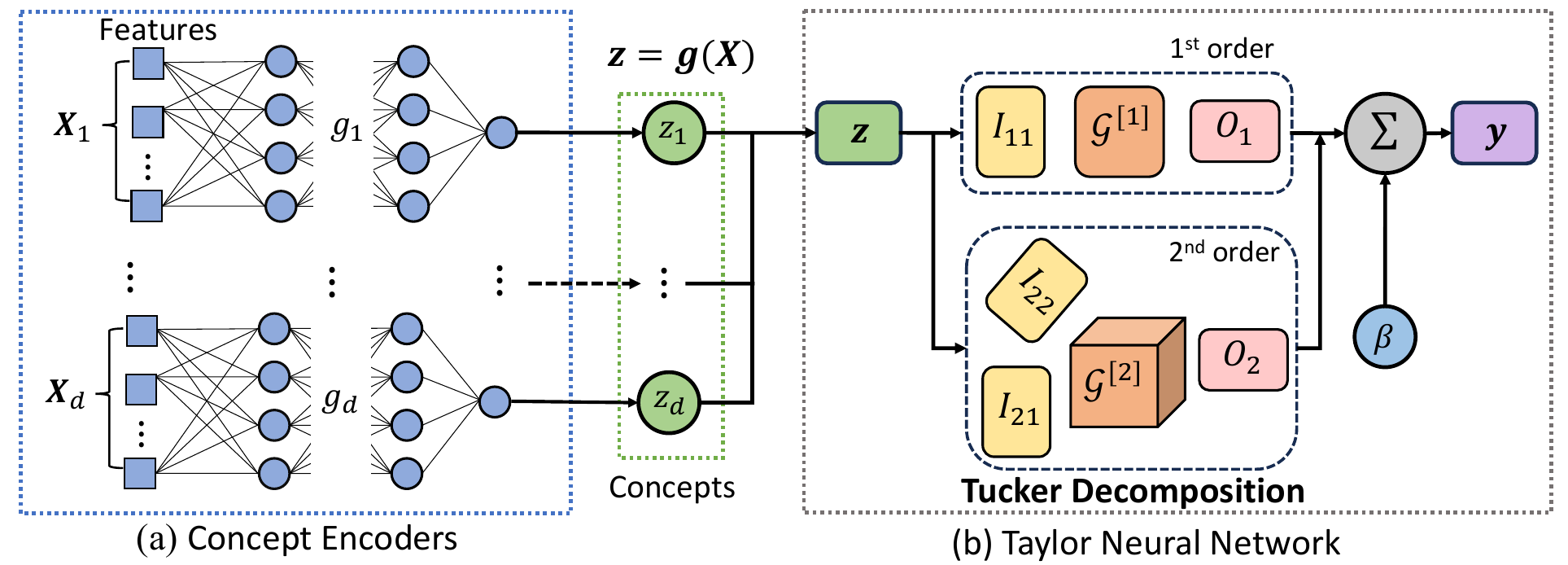}
    \caption{The Overall framework of CAT. It consists of two main components: concept encoders and Taylor Neural Networks (TaylorNet). Each concept encoder embeds a group of low-level features into a one-dimensional high-level concept representation. The TaylorNet is a white-box model that uses the high-level concept representations to make predictions.}
    \label{fig:ncm}
\vspace{-1em}
\end{figure*}

To overcome these limitations, we propose a novel interpretable concept-based Taylor additive model, called CAT, that can explain predictions using high-level concepts without relying heavily on domain experts to label concepts and their ground-truth values. As shown in Fig.~\ref{fig:ncm}, the proposed CAT consists of two main components: (i) concept encoders and (ii) a white-box Taylor Neural Network (TaylorNet). Specifically, the concept encoders aim to learn high-level concept representations from low-level features, where each encoder produces a one-dimensional representation from a cluster of features. TaylorNet aims to approximate non-linear functions using polynomials without activation functions. It directly learns the relationship between the input and output with polynomials, largely improving the interpretability. A significant challenge is to reduce the computational complexity of TaylorNet with high-order polynomials. To overcome this, we adopt Tucker decomposition~\cite{tucker1966some,kolda2009tensor} to decompose the higher-order coefficients in Taylor expansion into a set of low-rank tensors.

We assess the proposed CAT across multiple benchmark datasets for tabular and visual reasoning tasks. The evaluation results demonstrate the good performance of our method on these datasets. It can achieve higher or comparable accuracy to the best baseline with a reduction in the number of required model parameters. Furthermore, we demonstrated the efficacy of our concept encoder by integrating it with other interpretable baselines. Importantly, CAT offers improved explanation capabilities by articulating model predictions through human-understandable concepts.

In summary, the main contributions of this work include: (1) We introduce CAT, a novel interpretable framework capable of explaining DNNs predictions through high-level concepts; (2) We develop a white-box TaylorNet to directly learn the relationships between the input and output with polynomials, largely enhancing interpretability; (3) Extensive experimental results demonstrate that our method outperforms or competes with the baselines on six benchmark datasets, achieving this  with fewer or a similar number of model parameters; and (4) We present a case study illustrating how CAT allow users to comprehend model predictions by categorizing input features into concepts using basic data understanding and describing predictions in terms of Taylor polynomial.

\section{Related Works}
\noindent
\textbf{Classical Interpretable Methods.} Early works on explaining black-box machine learning models mainly adopted perturbation-based post-hoc approaches~\cite{ribeiro2016model, lundberg2017unified} to estimate the feature importance. One typical example of these methods is LIME, which explains individual predictions of a neural network by approximating it with interpretable models, such as linear models or decision trees fitted over simulated data points by randomly perturbing the given inputs~\cite{ribeiro2016model}. However, recent research~\cite{ghorbani2019interpretation, slack2020fooling} showed that post-hoc approaches could be unfaithful to the predictions of the original model, and are computationally expensive~\cite{slack2021reliable}.

\noindent\textbf{Generalized Additive models (GAMs).} Some recent works have focused on a new line of interpretable machine learning methods, called Generalized Additive Models (GAMs). The basic idea of GAMs is to learn non-linear shape function of each input feature using a separate DNN and then use a linear combination of these shape functions to predict results. Since each input feature learned by DNN is independent, it can help us explain the predictions based on the corresponding shape function. For instance, some representative models, such as Neural Additive Models (NAM)~\cite{agarwal2021neural} and its tree-based variant NODE-GAM~\cite{chang2021node}, learn feature-wise shape functions using a separate DNN or oblivious decision tree ensembles for each individual feature. However, these methods requires a large number of parameters and are easy to suffer from overfitting. To overcome this challenge, some researchers introduced Neural Basis Models (NBM)~\cite{radenovic2022neural}, which use a single DNN to learn shared bases and then input the bases into one linear model for each feature to learn its shape function. By doing this, NBM can improve its scalability to many input features. More recently, Scalable Polynomial Additive Models (SPAM) was proposed to incorporate high-order feature interactions for prediction and explanation~\cite{dubey2022scalable}. Another recent study adopted soft decision trees with hierarchical constraints to learn sparse pairwise interactions, improving the scalability and interpretability of tree-based GAMs~\cite{ibrahim2023grand}. However, these approaches try to explain the predictions from low-level features instead of high-level concepts that humans can easily understand. In particular, when the number of input features is large, it is still hard to fully interpret the prediction results.

\noindent\textbf{Concept-based Interpretability.} Our proposed method is closely related to concept-based learning, where models learn intermediate mappings from raw inputs to human-specified high-level concepts, then use only these annotated concepts to make final predictions. Recently, concept-based models have emerged as an important interpretable machine learning framework for many applications, such as medical diagnosis~\cite{de2018clinically, graziani2018regression, clough2019global, koh2020concept}, visual question answering~\cite{yi2018neural, bucher2019semantic}, and image recognition~\cite{losch2019interpretability, chen2020concept, koh2020concept}. For example, Koh et al. ~\cite{koh2020concept} proposed Concept Bottleneck Model (CBM) to predict concepts annotated by medical experts from X-ray image data. However, this approach may result in low accuracy when concept labels do not contain all the necessary information for downstream tasks~\cite{koh2020concept}. To deal with this problem, Mahinpei et al.~\cite{mahinpei2021promises} proposed to use an additional set of unsupervised concepts to improve the accuracy at the cost of the interpretability. Then, some researchers tried to trade-off the interpretability and accuracy in a follow-up work~\cite{zarlenga2022concept, havasi2022addressing}. However, existing work are heavily dependent on the human annotated concepts with values, which is infeasible in many real-world settings. Different from prior works, we propose to group input variables into high-level concepts based on categories without imposing specific values on them, thereby reducing the cost of human annotations and making it more practical and flexible to real-world applications.

\section{Preliminaries}
In this section, we first describe the research problem, and then review the basic knowledge of Taylor series expansion and Tucker decomposition.

\subsection{Problem Definition}
Given a multivariate input data $\vect{X}$, our goal is to learn a prediction model defined by function $\vect{h}:\vect{X}\rightarrow \vect{y}$ that maps $\vect{X}$ to the vector $\vect{y}$ of target labels representative of a regression or classification problem. In this work, we consider the problem of creating an interpretable model that can explain its predictions based on abstractions of the input features known as concepts. To this end, we assume that the data come with some descriptive information or metadata about the input features, so that we can manually and/or empirically divide the input space $\vect{X}$ into $d$ groups of features $\{\vect{X}_1,\vect{X}_2,\cdots,\vect{X}_d\}$ representing high-level concepts. In order to condition the prediction of target variable $\vect{y}$ on high-level concepts, we decompose the original function $\vect{h}$ into 2 functions: concept encoders $\vect{g}$ and target predictor $\vect{f}$, such that $\vect{h}=\vect{f}\circ \vect{g}$. In particular, concept encoders $\vect{g}=\{g_m:X_m\rightarrow z_m|m=1,\cdots,d\}$ consist of $m$ encoders, each $g_m$ individually maps one feature group $\vect{X}_m$ to a 1-D \textit{scalar} representation $z_m$. Then, the concept representations are combined into an intermediate concept vector $\vect{z}=\{z_1,\dots,z_m\}$. Finally, target predictor $\vect{f}:\vect{z}\rightarrow \vect{y}$ uses concept vector $\vect{z}$ as input to predict the target $\vect{y}$. If the learned concepts $\vect{z}$ are semantically meaningful and the predictor $\vect{f}$ is interpretable, then humans can interpret the model's decision process by attributing its predictions to the relevant concepts. Table~\ref{tab:notations} summarizes the main notations that will be used throughout the paper.
\vspace{-1em}
\begin{table}[!htb]
    \caption{Summary of notations.}\label{tab:notations}
    \vspace{-0.5em}
    \centering
    \resizebox{.9\columnwidth}{!}{%
    \begin{tabularx}{\linewidth}{|p{2.7cm}|X|}
        \hline
        \textbf{Notation} & \textbf{Definition} \\ \hline
        $\vect{X}$ &  multivariate input matrix\\ 
        \hline
        $d$ & number of concept groups \\ 
        \hline
        $\vect{z}$ & concept embedding vector \\ 
        \hline
        $\vect{f}:\vect{z}\rightarrow \vect{y}$  & mapping function from concepts to target variable(s)\\ \hline
        $\vect{g}=\{g_m:X_m\rightarrow z_m$ $|\; m=1,\cdots,d\}$ & set of concept encoders, each operates on one group of features $\vect{X}_m$, $m=1,\dots,d$.\\ \hline
        $\vect{h}=\vect{f}\circ \vect{g}$ & function representing the entire framework, which is $\vect{f}$ composed with $\vect{g}$ in our method.\\ \hline
        $\otimes$, $\times_n$, $\bar{\times}_n$  &
        Kronecker product, mode-$n$ matrix product,  mode-$n$ vector product \\  \hline
        $k, N$ & term order of Taylor polynomial, the total order of Taylor polynomial\\  \hline
        $\Delta \bm{z}$ & Input of TaylorNet \\  \hline
        $\bm{\mathcal{G}}^{[k]}$ & Learnable core tensor of TaylorNet \\ 
        \hline
    \end{tabularx}%
}
\vspace{-1em}
\end{table}

\subsection{Taylor Polynomials}\label{subsect:taylor-poly}
In mathematics, Taylor's theorem~\cite{thomas2010thomas} states that given a vector-valued multivariate function $\bm{f}:\mathbb{R}^{d}\rightarrow\mathbb{R}^{o}$, its approximation using a Taylor polynomial of order $N$ at a point $\vect{z}=\vect{z}_0$ is given by: 
\begin{equation}
\vect{f}\left(\vect{z}\right)\approx\sum_{k=0}^{N}\frac{1}{k!}\left[\sum_{j=1}^d\left(\Delta z_j \frac{\partial}{\partial z_j}\right) \right]^k \vect{f}\bigg{|}_{\vect{z}_0},
\end{equation}
where $\vect{z}\in \mathbb{R}^{d}$, $\vect{x}_0 \in \mathbb{R}^{d}$, $\Delta z_j=z_j - z_{j,0}$, and $j=1,\cdots,d$. This Taylor polynomial can also be expressed as the followig tensor form~\cite{chrysos2020p}:
\begin{equation}
\label{eq:taylor}
\vect{f}\left(\vect{z}\right)\approx\vect{f}\left(\vect{z}_0\right)+\sum_{k=1}^{N} \left(\mathbfcal{W}^{[k]}\prod_{j=2}^{k+1}{\bar \times}_j \Delta \vect{z}^\top \right),
\end{equation}
where $\bar{\times}_n$ denotes the \textit{mode-$n$ matrix product}~\cite{kolda2009tensor}, $\Delta 
\vect{z}=\vect{z} - \vect{z}_0$, and parameter tensors $\mathbfcal{W}^{[k]}\in \mathbb{R}^{o\prod_{n=1}^k\times d}$ for $k=1,\cdots,N$ are the $k$-th order \textit{scaled derivatives} of $\vect{f}$ at point $\vect{z}=\vect{z}_0$. According to the Stone–Weierstrass theorem~\cite{stone1948generalized}, polynomials defined in Eq.~\ref{eq:taylor} can approximate any  continuous function defined in a closed interval as closely as desired. In general, Taylor polynomials of higher order produce more precise approximations. However, as the order increases, the computational complexity grows exponentially, leading to scalability issues in high-dimensional input data.

\subsection{Tucker Decomposition}
Tucker decomposition is a tensor decomposition technique~\cite{kolda2009tensor}, aiming to decompose a tensor into a set of factor matrices and a small core tensor~\cite{tucker1966some}. Fundamentally, Tucker decomposition can be considered as a generalization of principal component analysis to higher-order analysis. In particular, given an $N$-way tensor $\mathbfcal{W}$, the Tucker decomposition of $\mathbfcal{W}$ is given by:
\begin{equation}\small
\label{eq:tucker}
    \mathbfcal{W}=\mathbfcal{G}\times_1 \vect{U}^{(1)}\times_2 \vect{U}^{(2)}\times_3\cdots \times_N \vect{U}^{(N)},
\end{equation}
where $\mathbfcal{G}$ is the $N$-way core tensor, and $\vect{U}^{(k)}$ for $k=1,\cdots,N$ are the factor matrices along corresponding mode $k$. Some researchers~\cite{kolda2009tensor} alternatively express Eq.~\ref{eq:tucker} in matricized form using Kronecker products as:
\begin{equation}\small
\label{eq:tucker_kron}
\vect{W}_{(k)}=\vect{U}^{(k)}\vect{G}_{(k)}\left(\vect{U}^{(N)}\otimes\cdots \otimes\vect{U}^{(k+1)}\otimes\vect{U}^{(k-1)}\otimes\cdots \otimes\vect{U}^{(1)}\right)^\mathrm{T},
\end{equation}
where $\otimes$ denotes Kronecker product, and $\vect{W}_{(k)}$ and $\vect{G}_{(k)}$ are the mode-$k$ matricization of the tensors $\mathbfcal{W}$ and  $\mathbfcal{G}$ respectively.

\section{Methodology}
In this section, we first introduce the motivation behind learning high-level concepts from input features as the foundation of an interpretable machine learning framework. Given the learned concepts, we propose an interpretable TaylorNet, an expressive approximation algorithm, to capture concept semantics and interactions for building accurate prediction models, as illustrated in Fig.~\ref{fig:ncm}. To reduce the computational costs of TaylorNet, we adopt Tucker decomposition to decompose the higher-order coefficients in Taylor expansion into a set of low-rank tensors.

\subsection{Concept Encoders} \label{subsect:concept}
Generalized Additive Models (GAMs)~\cite{hastie2017generalized} is an emerging paradigm that can interpret machine learning models at a feature level. The intuition behind these models is similar to that of linear regression, which learns the approximation $\vect{h}(\vect{X})$ of dependent variable $\vect{y}$ by parameterizing a linear combination $\beta_0+\sum_{i=1}^n\beta_i X_i$
of input features $\vect{X}=X_1,\cdots,X_n$. In typical GAMs, each $\beta_i X_i$ is replaced by shape function $s_i(X_i)$ that transforms $X_i$ into a smooth representation, such that the sum of $s_i(X_i)$'s is the generalized smooth estimate of $\vect{y}$. Furthermore, GAMs can be extended to model pairwise and higher-order feature interactions~\cite{lou2012intelligible} for improved accuracy. In this case, $\vect{h}(\vect{X})$ is given by:
\begin{equation} \small
\label{eq:gam}
    \vect{h}(\vect{X})=\beta_0+\sum_{i=1}^n s_i(X_i)+\sum_{j\ne i} s_{ij}(X_i,X_j)+\cdots+s_{1\cdots n}(\vect{X}),
\end{equation}
where $s_{ij}(X_i,X_j)$ in the third term is the second-order or pairwise interaction between $X_i$ and $X_j$, and the subsequent terms denotes higher-order transformation of up to $n$-way feature interaction among all $X_i$'s. Evidently, as the order of feature interaction increases, high-order GAMs approach the level of expressiveness closer to that of fully-connected DNNs, leading to better performance. However, beyond third-order interactions, these models will be hardly interpretable~\cite{lou2012intelligible}, especially when the number of features $n$ is large. In addition, we observe that among all the input features and their high-order interactions, hardly all, or only a few of them are meaningful to the prediction accuracy and human interpretation. Therefore, in order to limit the degree of interaction between input variables to those that are relevant and intelligible to humans, we propose to learn high-level concepts from each group of semantically related features.

Given a multivariate input space $\vect{X}$, we aim to partition $\vect{X}$ into $d$ groups of features $\{\vect{X}_1,\vect{X}_2,\cdots,\vect{X}_d\}$ that represent the corresponding high-level concepts. For ease of explanation, we consider a general application of concept-based method on tabular data. Specifically, we assume that the studied datasets are accompanied by some form of metadata detailing the meaning of corresponding input features. Given the metadata, we can group closely related features into high-level concepts manually and/or in conjunction with possible analyses on empirical similarity and correlation of feature metadata or values~\cite{yao2022concept, kuzudisli2023review}. 
Since these concept groups convey high-level representations for their low-level features, we can disregard the less meaningful terms in Eq.~\ref{eq:gam} such that $\vect{h}(\vect{X})$ can be approximated as the sum of concepts representations and their high-order interactions:
\begin{equation}\footnotesize
\label{eq:concepts}
    \vect{h}(\vect{X})\approx\beta_0+\sum_{k=1}^d s_k(\vect{X}_k)+\sum_{l\ne k} s_{kl}(\vect{X}_k,\vect{X}_{l})+\cdots+s_{1\cdots d}(\vect{X}_1,\cdots,\vect{X}_d),
\end{equation}
where the first term is the sum of high-level concept representations for each group of closely related features, and each following term represents the interaction among groups of concepts. If the number of concepts $d$ is small, then the order of interaction within the model is much lower than in Eq.~\ref{eq:gam}, leading to a more interpretable model. Additionally, interactions among concepts can serve as the proxy for the interactions among enclosed features, allowing humans to interpret the interactions among a large number of features at an abstract level.

To learn the latent vector representation $\vect{z}$ of the high-level concepts from tabular data, we utilize an ensemble of $d$ DNN concept encoders $\vect{g}=\{g_m:X_m\rightarrow z_m|m=1,\cdots,d\}$, each operating on a group of features as illustrated in Fig.~\ref{fig:ncm}(a) to obtain intermediate concepts $\vect{z}=\vect{g}(\vect{X})$. For higher-dimensional input data such as 2D image, this is achieved by using specifically designed concept discovery algorithms such as disentangled representation learning~\cite{kim2018disentangling, shao2020controlvae, shao2022rethinking}, which can extract the disentangled latent factors $\vect{z}$ representing high-level visual concepts from images. Furthermore, we can substitute $\vect{f}(\vect{z})=\vect{f}(\vect{g}(\vect{X}))=\vect{h}(\vect{X})$ into Eq.~\ref{eq:concepts}. Then by defining each $k$th-order concept interaction as the product of $k$ corresponding concept embeddings and a weight parameter, $\vect{f}(\vect{z})$ can be expressed in polynomial form of order $N\leq d$ as follows:
\begin{equation}\small \label{eq:polynomial}
\vect{f}\left(\vect{z}\right)\approx\beta_0+\sum_{k=1}^{N} \left(\mathbfcal{W}^{[k]}\prod_{j=2}^{k+1}\bar{\times}_j \vect{z} \right),
\end{equation}
where parameter tensors $\mathbfcal{W}^{[k]}\in \mathbb{R}^{o\prod_{m=1}^k\times d}$ for $k=1,\cdots,N$ characterize the $k$-th order concept interactions. 

However, one major challenge is that the Taylor polynomial in Eq.~\ref{eq:polynomial} will be computationally expensive as its order is large. To deal with this problem, we propose to develop a novel TaylorNet based on Tucker decomposition, which allows each mode to have more possible interactions between the latent factors, so that it is more expressive than CP tensor decomposition in existing works~\cite{chrysos2020p, dubey2022scalable}.

\subsection{Learning Predictive Models with TaylorNet} \label{subsect:taylornet}
We aim to develop a Taylor Neural Network (TaylorNet) as given in Eq.~\ref{eq:taylor} to approximate function $\vect{f}$ with latent concept vector $\vect{z}$ as input. Since $\vect{f}$ is unknown at training time, we consider $\vect{f}(\vect{z}_0)$ and $\mathbfcal{W}^{[k]}$'s as \textit{learnable parameters}. As tensors $\mathbfcal{W}^{[k]}$ denote the  $k$-th order scaled derivatives of function $\bm{h}$, the number of parameters required to learn $\mathbfcal{W}^{[k]}$ grows exponentially with respect to polynomial order $k$ (i.e., $\mathbfcal{O}(d^k)$). To overcome this issue, we adopt Tucker decomposition on $\mathbfcal{W}^{[k]}$ following Eq.~\ref{eq:tucker} as follows:
\begin{equation}\label{eq:tucker_taylor}\small
\begin{split}
    \mathbfcal{W}^{[k]} & = \mathbfcal{G}^{[k]} \times_1 \vect{O}_k \times_2 \vect{I}_{k1} \dots \times_{k+1}  \vect{I}_{kk} \\
    & = \mathbfcal{G}^{[k]} \times_1 \vect{O}_k \prod_{j=1}^k \times_{j+1} \vect{I}_{kj},
\end{split}
\end{equation}
where $\mathbfcal{G}^{[k]} \in \mathbb{R}^{r_{out,k} \prod_{j=1}^k \times r_{in,k,j}}$ is the core tensor; $\vect{I}_{kj} \in \mathbb{R}^{d \times r_{in,k,j}}$ and $\vect{O}_k \in \mathbb{R}^{o \times r_{out,k}}$ for $j = 1, \dots, k$ are input and output factor matrices respectively. For each $k$-th-order term of the Taylor polynomial, $r_{in,k,j}$ and $r_{out,k}$ are denoted as the ranks of Tucker decomposition corresponding to the $j$-th input and output dimension. 

We further substitute Eq.~\ref{eq:tucker_taylor} into Eq.~\ref{eq:taylor}, such that the $k$-th term of Taylor polynomial can be written as:
\begin{equation} \small \label{eq:taylor_term}
\mathbfcal{W}^{[k]} \prod_{j=2}^{k+1} \times_j \Delta \vect{z}^\top = \mathbfcal{G}^{[k]} \times_1 \vect{O}_k \left(\prod_{i=1}^k \times_{i+1} \vect{I}_{ki}\right) \left(\prod_{j=1}^{k} \times_{j+1} \Delta \vect{z}^\top\right)
\end{equation}

Subsequently, we apply the commutative and associative properties of mode-$n$ product~\cite{kolda2009tensor} on Eq.~\ref{eq:taylor_term} yielding:
\begin{equation} \label{eq:tucker_taylor2}\small
\begin{split}
\mathbfcal{W}^{[k]} \prod_{j=2}^{k+1} \times_j \Delta \vect{z}^\top
&= \mathbfcal{G}^{[k]} \times_1 \vect{O}_k \left[\prod_{j=1}^k \times_{j+1} \left(\Delta\vect{z}^\top \vect{I}_{kj}\right)\right] \in \mathbb{R}^{o}.
\end{split}
\end{equation}

Since current deep learning frameworks such as Pytorch and Tensorflow do not support batch-wise mode-$n$ multiplication, we additionally follow the mode-$n$ unfolding rule as previously demonstrated by Eq.~\ref{eq:tucker} and Eq.~\ref{eq:tucker_kron} to rewrite Eq.~\ref{eq:tucker_taylor2} using Kronecker products for ease of implementation. We obtain the following:
\begin{equation}\label{eq:final_taylor_term}\small
    \mathbfcal{W}^{[k]} \prod_{j=2}^{k+1} \times_j \Delta \vect{z}^\top = \vect{O}_k \vect{G}_{k} \left[ \left(\vect{I}_{kk}^\top\Delta \vect{z}\right) \otimes \dots \otimes \left(\vect{I}_{k1}^\top\Delta \vect{z}\right)\right],
\end{equation}
where $\vect{G}_{k} = \vect{{G}}^{[k]}_{(1)}$ denotes the mode-$1$ matricization of tensor $\bm{\mathcal{G}}^{[k]}$.

Afterwards, the remaining step is to substitute the above Eq.~\ref{eq:final_taylor_term} into the original Taylor polynomial in Eq.~\ref{eq:taylor}, which is given by:
\begin{equation}\label{eq:taylorNet}\small
    \vect{f}(\vect{z}) = \vect{\beta} + \sum_{k=1}^{N} \vect{O}_k \vect{G}_{k} \left[ \left(\vect{I}_{kk}^\top\Delta \vect{z}\right) \otimes \dots \otimes \left(\vect{I}_{k1}^\top\Delta \vect{z}\right)\right],
\end{equation}

\noindent where $\vect{\beta}=\vect{f}(\vect{z}_0)$, $\vect{O}_k$, $\vect{G}_{k}$, and $\vect{I}_{kj}^\top~(k = 1, \dots, N;~j = 1, \dots, k)$ are learnable parameters. Fig.~\ref{fig:ncm}(b) visualizes the proposed TaylorNet with Tucker decomposition. In theory, it is possible to stack multiple layers of TaylorNet to construct high-order polynomials with high expressivity. However, to reduce the number of model parameters and allow straightforward interpretation of the model's prediction, we only use single-layer Taylor network with small orders of the Taylor polynomial (e.g., 2 or 3).

\noindent\textbf{Computational Complexity of TaylorNet.} Given $d$ and $o$ as the input and output dimension respectively, the original computational complexity of the $k$-order term of Taylor polynomial is $\mathbfcal{O}(od^k)$. Using Tucker decomposition results in a substantial reduction in the number of parameters to $\mathbfcal{O}\left(r_{out,k}\prod_{j=1}^k+or_{out,k}+d\sum_{j=1}^k r_{in,j,k}\right)$. In particular, when the rank of the core tensor in Tucker decomposition is much smaller than $d$ and $o$, the training time of TaylorNet will be improved by orders of magnitude.
\section{Experiments}
In this section, we conduct extensive experiments to evaluate the performance of the proposed CAT across multiple tabular and image benchmark datasets. Additionally, we do a case study to demonstrate that our method can effectively explain the prediction results using high-level concepts. Finally, we also demonstrate the effectiveness of the concept encoders applied to other baselines.

\subsection{Datasets}
We conduct experiments on six real-world benchmarks, including four tabular datasets (1-4) and two image datasets (5-6), as below:
\begin{enumerate}
    \item \textbf{Airbnb Listings and Reviews} (Airbnb~\cite{airbnb}): 
    This dataset encompasses over 250,000 Airbnb listings across major global cities. We formulate a regression task on this dataset, predicting the listing price based on host details, locations, and property information. With 126 features, we manually categorize them into 6 concepts.
    \item \textbf{WiDS Diabetes Detection} (Diabetes~\cite{diabetes}): This binary classification dataset indicates whether a patient is diagnosed with diabetes within the initial 24 hours of admission. Features include demographic information, medical history, and various lab metrics, totaling 176 features grouped into 6 categories by data providers~\cite{diabetes}.
    \item \textbf{COMPAS Recidivism} (COMPAS~\cite{larson2016propublica}): This binary classification dataset aims to predict the risk of repeated offense among convicted criminals. It includes 6 input features: age, sex, race, priors count, charge degree, and custody length. Based on their semantics, we divide these features into 2 groups: demographic (first 3) and criminal history (last 3).
    \item \textbf{Daily and Sports UCI-HAR} (UCI-HAR~\cite{misc_human_activity_recognition_using_smartphones_240}): This multi-class dataset from the UCI ML Repository involves predicting 6 daily activities performed by volunteers over a period of time while wearing a smartphone with multiple sensors on the waist. The signal sequences come from 3 sensor types: body accelerometer, gravity accelerometer, and gyroscope. Bulbul et al.~\cite{misc_human_activity_recognition_using_smartphones_240} further derived jerk signals from accelerometers, then calculated triaxial signal magnitudes and fast Fourier transform frequencies on them to obtain a total of 18 different signal types. To obtain tabular features from time series data, they extracted 33 features including descriptive statistics, energy, and autocorrelation coefficients from each signal type to gather 561 features in total for the training data. We consider these 18 signal types as high-level concepts in our experiment.
    \item \textbf{MNIST}~\cite{lecun1998mnist}: This dataset consists of 70,000 images of handwritten digits (0-9) for multi-class classification. Following previous works~\cite{kim2018disentangling, shao2022rethinking}, we utilize disentangled representation learning to extract 6 high-level latent factors like style and shape from MNIST images as high-level-concepts for performing classification.
    \item \textbf{CelebA}~\cite{lee2020maskgan}: This dataset contains 30,000 high-quality RBG images of center-aligned facial photographs of celebrities. We formulate a binary classification task on this dataset based on the gender annotations (male/female) provided by Lee et al.~\cite{lee2020maskgan}. Additionally, we extract 9 high-level concepts such as skin tone, hair azimuth, hair length, etc. from facial images using disentangled representation learning~\cite{shao2022rethinking}. 
\end{enumerate}

The summary of these datasets is presented in Table~\ref{tab:data}. Furthermore, we adopt a fixed random sample ratio of 80-10-10 for training, validation, and testing for Airbnb, COMPAS, and Diabetes. Since the train-test splits are provided in UCI-HAR, MNIST, and CelebA datasets, we reserve 10\% of each training dataset for validation.

\begin{table}[!t]
\caption{Summary of Experimental Datasets}
\vspace{-0.1in}
\label{tab:data}
\centering
    \resizebox{\columnwidth}{!}{%
\begin{tabular}{ccccccc}
\hline\hline
Name       & Airbnb & COMPAS & Diabetes & UCI-HAR & MNIST & CelebA\\ \hline
Instances    & 275,598 & 6,172 & 130,157  & 10,299  & 70,000  & 30,000   \\
Features     & 126 & 6    & 176      &  561  & - & -   \\
Concepts     & 6  & 2     & 6        & 18   & 6 & 9      \\
Classes     & -       & 2   & 2     & 6  & 10  & 2      \\
Feature Type & Mixed & Mixed  & Mixed    & Numeric  & Image  & Image  \\ \hline\hline
\end{tabular}%
}
\vspace{-1em}
\end{table}

\subsection{Baselines}
We compare the performance of CAT with the following baselines for both classification and regression problems. Note that the following two methods, MLP and XGBoost, are uninterpretable while the remaining are interpretable.
\begin{itemize}
\item \textbf{Multi-layer Perceptron (MLP)}: MLP serves as a standard uninterpretable black-box neural network, setting the upper bound on prediction performance.
\item \textbf{Gradient Boosted Trees (XGBoost)}: XGBoost~\cite{chen2016xgboost} is  a robust machine learning algorithm based on an ensemble of decision trees. Given the typically large number of trees, the model becomes uninterpretable. We utilize the \texttt{xgboost} library in our experiments.

\item \textbf{Explainable Boosting Machines (EBM)}: EBM models are Generalized Additive Models (GAMs) that leverage millions of shallow bagged trees to learn a shape function for each feature individually~\cite{lou2012intelligible, caruana2015intelligible}. We implement EBM using the \texttt{interpretml} library.
\item \textbf{Neural Additive Models (NAM)}:  NAMs~\cite{agarwal2021neural} extend GAMs with neural network components to learn one Multi-Layer Perceptron (MLP) per feature. We implement the original NAM following prior work~\cite{dubey2022scalable} to expedite training time.
\item \textbf{Neural Basis Models (NBM)}: As an extension of NAMs, NBMs learn a set of basis functions shared across all features instead of individual shape functions for each feature~\cite{radenovic2022neural}.
\item \textbf{Scalable Polynomial Additive Models (SPAM)}: Representing the current state-of-the-art GAMs, SPAMs incorporate high-order interactions between shape functions learned by prior NAMs using polynomial neural networks~\cite{dubey2022scalable}.
\item \textbf{Grand-Slamin Additive Modeling with Structural Constraints (Grand-Slamin)}: This method utilizes tree-based GAMs with sparsity and structural constraints to selectively learn second-order interactions between shape functions derived from soft decision trees~\cite{irsoy2012soft} for enhanced interpretability and scalability~\cite{ibrahim2023grand}. 
\end{itemize}
Note that we do not compare with concepts-based interpretability methods~\cite{koh2020concept, chen2020concept, zarlenga2022concept}, since these approaches require domain experts to label a lot of concepts and then specify the corresponding ground-truth values. In addition, the above interpretable baselines, such as EBM, NAM, NBM, SPAM, and Grand-Slamin are difficult to interpret if number of features are large, since they treat each feature individually without organizing them into concepts; they are also computationally expensive because they use large trees or DNNs on each feature.

\subsection{Experimental Settings}

\textbf{CAT Architecture:} We employ the following structure for the concept encoders: a Multi-Layer Perceptron (MLP) with 3 hidden layers having 64, 64, and 32 hidden units, along with LeakyReLU activation~\cite{xu2020reluplex}, following recent approaches in concept-based methods~\cite{koh2020concept, zarlenga2022concept}. For the TaylorNet, we opt for rank $r=8$ for Taylor order 2 and $r=16$ for Taylor order 3. The initial value of Taylor series expansion is 0. In Section~\ref{subsect:ablation}, we will investigate the impact of these hyperparameters on prediction performance.

\noindent \textbf{Implementation Details.} MLP, NAM, NBM, SPAM, Grand-Slamin, and CAT models are implemented in Pytorch and trained using the Adam optimizer with decoupled weight decay regularization (AdamW~\cite{loshchilov2017decoupled}) on A6000 GPU machines with 48GB memory. We train the model with 100 epochs on all six datasets with early stopping. For CAT on all datasets, we tune the starting learning rate in the interval $[0.0001, 0.1]$, concept encoder dropout and TaylorNet dropout coefficients in the discrete set $\{0, 0.05, 0.1, 0.2, 0.3, 0.4, 0.5\}$. The best-performing hyperparameters are determined using the validation set through grid search. A similar tuning procedure is applied to MLP, NAM, NBM, SPAM, and Grand-Slamin. Finally, for EBMs and XGBoost, we use CPU machines and follow the training guidelines provided by corresponding code libraries.

\noindent \textbf{Evaluation Details.} We report the appropriate performance metrics for different prediction tasks: (i) Root Mean-Squared Error (RMSE) for regression task; and (ii) Accuracy and Macro-F1 for classification tasks. Moreover, we report the average performance over 3 runs with different random seeds.

\begin{table*}[!t]
\caption{Performance comparison between CAT and prior ML methods on benchmark datasets averaged over three random seeds. Note that the best result is highlighted in bold black and the second best is highlighted in {\color[HTML]{187F1A} green}. We can observe that CATs generally outperform NAM, NBM, SPAMs, and Grand-Slamin, and outperform EBM in five out of six benchmarks.}
\label{tab:main-results}
\vspace{-0.5em}
\begin{adjustbox}{width=\textwidth}\small
\begin{tabular}{|lrrrrrrrrrrr|}
\hline
\multicolumn{1}{|c|}{\multirow{2}{*}{\textbf{Models}}} & \multicolumn{1}{c|}{\textbf{AirBnb}} & \multicolumn{2}{c|}{\textbf{COMPAS}}                         & \multicolumn{2}{c|}{\textbf{Diabetes}}                                      & \multicolumn{2}{c|}{\textbf{UCI-HAR}}                                    & \multicolumn{2}{c|}{\textbf{MNIST}}                                         & \multicolumn{2}{c|}{\textbf{CelebA}}                                  \\ \cline{2-12} 
\multicolumn{1}{|c|}{}                                 & \multicolumn{1}{c|}{RMSE↓}           & \multicolumn{1}{c|}{Acc↑}   & \multicolumn{1}{c|}{Macro-F1↑} & \multicolumn{1}{c|}{Acc↑}            & \multicolumn{1}{c|}{Macro-F1↑}       & \multicolumn{1}{c|}{Acc↑}            & \multicolumn{1}{c|}{Macro-F1↑}       & \multicolumn{1}{c|}{Acc↑}            & \multicolumn{1}{c|}{Macro-F1↑}       & \multicolumn{1}{c|}{Acc↑}            & \multicolumn{1}{c|}{Macro-F1↑} \\ \hline
\multicolumn{12}{|c|}{\textbf{Black-box Baselines}}                                      \\ \hline
\multicolumn{1}{|l|}{XGBoost}                          & \multicolumn{1}{r|}{0.5131}          & \multicolumn{1}{r|}{0.6713}       & \multicolumn{1}{r|}{0.6664}          & \multicolumn{1}{r|}{0.8258}          & \multicolumn{1}{r|}{0.7129}          & \multicolumn{1}{r|}{0.9796}          & \multicolumn{1}{r|}{0.9381}          & \multicolumn{1}{r|}{0.9903}                & \multicolumn{1}{r|}{0.9510}                & \multicolumn{1}{r|}{0.7734}                &   0.7528                             \\ \hline
\multicolumn{1}{|l|}{MLP}                              & \multicolumn{1}{r|}{0.5437}          & \multicolumn{1}{r|}{0.6599}       & \multicolumn{1}{r|}{0.6468}          & \multicolumn{1}{r|}{0.8257}          & \multicolumn{1}{r|}{0.7232}          & \multicolumn{1}{r|}{0.9840}          & \multicolumn{1}{r|}{0.9514}          & \multicolumn{1}{r|}{0.9902}                & \multicolumn{1}{r|}{0.9510}                & \multicolumn{1}{r|}{0.7768}                &   0.7601                             \\ \hline
\multicolumn{12}{|c|}{\textbf{Interpretable Models}}                                         \\ \hline
\multicolumn{1}{|l|}{EBM}                              & \multicolumn{1}{r|}{0.6344}          & \multicolumn{1}{r|}{0.6710} & \multicolumn{1}{r|}{0.6643}    & \multicolumn{1}{r|}{0.8269}          & \multicolumn{1}{r|}{0.7031}          & \multicolumn{1}{r|}{\textbf{0.9867}}          & \multicolumn{1}{r|}{\textbf{0.9593}}          & \multicolumn{1}{r|}{0.9808}                & \multicolumn{1}{r|}{0.9030}                & \multicolumn{1}{r|}{0.7413}                &    0.7061                            \\ \hline
\multicolumn{1}{|l|}{NAM}                              & \multicolumn{1}{r|}{0.6681}          & \multicolumn{1}{r|}{0.6699} & \multicolumn{1}{r|}{0.6623}    & \multicolumn{1}{r|}{0.8242}          & \multicolumn{1}{r|}{0.7199}          & \multicolumn{1}{r|}{0.9785}          & \multicolumn{1}{r|}{0.9346}          & \multicolumn{1}{r|}{0.9723}          & \multicolumn{1}{r|}{0.8635}          & \multicolumn{1}{r|}{0.7441}          & 0.7226                         \\ \hline
\multicolumn{1}{|l|}{NBM}                              & \multicolumn{1}{r|}{0.6637}          & \multicolumn{1}{r|}{0.6742} & \multicolumn{1}{r|}{0.6708}    & \multicolumn{1}{r|}{0.8257}          & \multicolumn{1}{r|}{0.7167}          & \multicolumn{1}{r|}{0.9792}          & \multicolumn{1}{r|}{0.9367}          & \multicolumn{1}{r|}{0.9770}          & \multicolumn{1}{r|}{0.8878}          & \multicolumn{1}{r|}{0.7455}          & 0.7231                         \\ \hline
\multicolumn{1}{|l|}{SPAM (Order 2)}                   & \multicolumn{1}{r|}{0.5664}          & \multicolumn{1}{r|}{0.6659} & \multicolumn{1}{r|}{0.6569}    & \multicolumn{1}{r|}{0.8230}          & \multicolumn{1}{r|}{0.7242}          & \multicolumn{1}{r|}{0.9809}          & \multicolumn{1}{r|}{0.9414}          & \multicolumn{1}{r|}{0.9860}          & \multicolumn{1}{r|}{0.9318}          & \multicolumn{1}{r|}{0.7468}          & 0.7129                         \\ \hline
\multicolumn{1}{|l|}{SPAM (Order 3)}                   & \multicolumn{1}{r|}{0.5560}          & \multicolumn{1}{r|}{0.6688} & \multicolumn{1}{r|}{0.6608}    & \multicolumn{1}{r|}{0.8272}          & \multicolumn{1}{r|}{0.7188}          & \multicolumn{1}{r|}{0.9801}          & \multicolumn{1}{r|}{0.9388}          & \multicolumn{1}{r|}{0.9883}          & \multicolumn{1}{r|}{0.9426}          & \multicolumn{1}{r|}{{\color[HTML]{187F1A} \textbf{0.7642}}}          & 0.7385                         \\ \hline
\multicolumn{1}{|l|}{Grand-Slamin}                              & \multicolumn{1}{r|}{0.5811}          & \multicolumn{1}{r|}{0.6704} & \multicolumn{1}{r|}{0.6630}    & \multicolumn{1}{r|}{0.8266}          & \multicolumn{1}{r|}{0.7260}          & \multicolumn{1}{r|}{0.9800}          & \multicolumn{1}{r|}{0.9392}          & \multicolumn{1}{r|}{0.9864}          & \multicolumn{1}{r|}{0.9317}          & \multicolumn{1}{r|}{0.7537}          & 0.7241                         \\ \hline
\multicolumn{1}{|l|}{CAT (Order 2)}                    & \multicolumn{1}{r|}{{\color[HTML]{187F1A} \textbf{0.5486}}}          & \multicolumn{1}{r|}{{\color[HTML]{187F1A} \textbf{0.6772}}} & \multicolumn{1}{r|}{{\color[HTML]{187F1A} \textbf{0.6710}}}    & \multicolumn{1}{r|}{{\color[HTML]{187F1A} \textbf{0.8286}}}          & \multicolumn{1}{r|}{{\color[HTML]{187F1A} \textbf{0.7269}}}          & \multicolumn{1}{r|}{0.9814}          & \multicolumn{1}{r|}{0.9431}          & \multicolumn{1}{r|}{{\color[HTML]{187F1A} \textbf{0.9892}}}          & \multicolumn{1}{r|}{{\color[HTML]{187F1A} \textbf{0.9469}}}          & \multicolumn{1}{r|}{0.7609}          & {\color[HTML]{187F1A} \textbf{0.7436}}                        \\ \hline
\multicolumn{1}{|l|}{CAT (Order 3)}                    & \multicolumn{1}{r|}{\textbf{0.5461}} & \multicolumn{1}{r|}{\textbf{0.6793}} & \multicolumn{1}{r|}{\textbf{0.6726}}    & \multicolumn{1}{r|}{\textbf{0.8295}} & \multicolumn{1}{r|}{\textbf{0.7270}} & \multicolumn{1}{r|}{{\color[HTML]{187F1A} \textbf{0.9829}}} & \multicolumn{1}{r|}{{\color[HTML]{187F1A} \textbf{0.9480}}} & \multicolumn{1}{r|}{\textbf{0.9902}} & \multicolumn{1}{r|}{\textbf{0.9517}} & \multicolumn{1}{r|}{\textbf{0.7728}} & \textbf{0.7579}                \\ \hline
\end{tabular}
\end{adjustbox}
\end{table*}

\begin{table*}[!th]
\caption{Benchmarks on the number of parameters and training throughput (training examples per second) between interpretable ML methods on six datasets. Here the best result is highlighted in bold black and the second best is highlighted in {\color[HTML]{187F1A} green}. We omit the training throughput of EBMs since they are trained on CPU machines. Overall, CATs have lower number of parameters and higher training throughput than NAM, NBM, and SPAM. Despite having the most compact models for Airbnb and Diabetes, the number of parameters in EBM drastically grow when applied on larger datasets.}
\label{tab:params}
\vspace{-0.5em}
\begin{adjustbox}{width=\textwidth}\small
\begin{tabular}{|l|rrrrrr||rrrrrr|}
\hline
\multicolumn{1}{|c|}{}  & \multicolumn{6}{c||}{\textbf{Number of Parameters↓}}   & \multicolumn{6}{c|}{\textbf{Training Throughout (samples/sec)↑}}                                  \\ \cline{2-13} 
\multicolumn{1}{|c|}{\multirow{-2}{*}{\textbf{Models}}} & \multicolumn{1}{c|}{\textbf{Airbnb}}                        & \multicolumn{1}{c|}{\textbf{COMPAS}} & \multicolumn{1}{c|}{\textbf{Diabetes}}                      & \multicolumn{1}{c|}{\textbf{UCI-HAR}}                        & \multicolumn{1}{c|}{\textbf{MNIST}}                        & \multicolumn{1}{c||}{\textbf{CelebA}}   & \multicolumn{1}{c|}{\textbf{Airbnb}}                       & \multicolumn{1}{c|}{\textbf{COMPAS}}                        & \multicolumn{1}{c|}{\textbf{Diabetes}}                      & \multicolumn{1}{c|}{\textbf{UCI-HAR}}                      & \multicolumn{1}{c|}{\textbf{MNIST}}                         & \multicolumn{1}{l|}{\textbf{CelebA}} \\ \hline
EBM                                                     & \multicolumn{1}{r|}{ \textbf{28,110}} & \multicolumn{1}{r|}{{\color[HTML]{009901} \textbf{9,827}}}                & \multicolumn{1}{r|}{351,165}                                & \multicolumn{1}{r|}{12,661,020}                              & \multicolumn{1}{r|}{86,800}                                & 112,640                  & \multicolumn{1}{r|}{\_}                                    & \multicolumn{1}{r|}{\_}                                     & \multicolumn{1}{r|}{\_}                                     & \multicolumn{1}{r|}{\_}                                    & \multicolumn{1}{r|}{\_}                                     & \_                                   \\ \hline
NAM                                                     & \multicolumn{1}{r|}{822,780}                                & \multicolumn{1}{r|}{40,326}          & \multicolumn{1}{r|}{1,142,750}                              & \multicolumn{1}{r|}{2,398,500}                               & \multicolumn{1}{r|}{40,326}                                & 67,210                                 & \multicolumn{1}{r|}{9,553}                                 & \multicolumn{1}{r|}{4,929}                                  & \multicolumn{1}{r|}{6,793}                                  & \multicolumn{1}{r|}{1,667}                                 & \multicolumn{1}{r|}{46,684}                                 &     19,339                                 \\ \hline
NBM                                                     & \multicolumn{1}{r|}{76,897}                                 & \multicolumn{1}{r|}{64,664}          & \multicolumn{1}{r|}{82,071}                                 & \multicolumn{1}{r|}{\textbf{124,077}}                        & \multicolumn{1}{r|}{64,720}                                & 65,076                                 & \multicolumn{1}{r|}{24,792}                                 & \multicolumn{1}{r|}{5,059}                                 & \multicolumn{1}{r|}{17,308}                                 & \multicolumn{1}{r|}{3,917}                                 & \multicolumn{1}{r|}{51,247}                                 &    21,244                                  \\ \hline
SPAM (Order 2)                                          & \multicolumn{1}{r|}{1,719,219}                              & \multicolumn{1}{r|}{82,254}          & \multicolumn{1}{r|}{2,338,102}                              & \multicolumn{1}{r|}{7,657,734}                               & \multicolumn{1}{r|}{83,862}                                & 136,822                                & \multicolumn{1}{r|}{4,927}                                   & \multicolumn{1}{r|}{4,641}                                  & \multicolumn{1}{r|}{3,428}                                  & \multicolumn{1}{r|}{949}                                      & \multicolumn{1}{r|}{33,344}                                       &   17,324                                   \\ \hline
SPAM (Order 3)                                          & \multicolumn{1}{r|}{2,604,292}                              & \multicolumn{1}{r|}{124,982}         & \multicolumn{1}{r|}{3,617,729}                              & \multicolumn{1}{r|}{11,601,687}                              & \multicolumn{1}{r|}{128,998}                               & 207,634                                & \multicolumn{1}{r|}{3,259}                                   & \multicolumn{1}{r|}{4,544}                                  & \multicolumn{1}{r|}{2,271}                                  & \multicolumn{1}{r|}{646}                                      & \multicolumn{1}{r|}{28,608}                                &       15,090                                \\ \hline
Grand-Slamin                                                     & \multicolumn{1}{r|}{494,173}                                & \multicolumn{1}{r|}{\textbf{4,852}}          & \multicolumn{1}{r|}{952,176}                              & \multicolumn{1}{r|}{2,911,253}                           & \multicolumn{1}{r|}{16,984}                                &        13,044                          & \multicolumn{1}{r|}{2,091}                                 & \multicolumn{1}{r|}{4,725}                                  & \multicolumn{1}{r|}{1,316}                                  & \multicolumn{1}{r|}{909}                                 & \multicolumn{1}{r|}{48,157}                                 &    20,051                                  \\ \hline
CAT (Order 2)                                           & \multicolumn{1}{r|}{{\color[HTML]{009901} \textbf{48,742}}}                                 & \multicolumn{1}{r|}{14,354}          & \multicolumn{1}{r|}{\textbf{51,310}} & \multicolumn{1}{r|}{{\color[HTML]{009901} \textbf{161,138}}} & \multicolumn{1}{r|}{\textbf{880}}                          & \textbf{4,896}                         & \multicolumn{1}{r|}{ \textbf{90,333}} & \multicolumn{1}{r|}{\textbf{5,415}} & \multicolumn{1}{r|}{\textbf{59,958}} & \multicolumn{1}{r|}{\textbf{7,598}} & \multicolumn{1}{r|}{ \textbf{67,097}} &  \textbf{25,773}                                    \\ \hline
CAT (Order 3)                                           & \multicolumn{1}{r|}{52,990}                                 & \multicolumn{1}{r|}{18,514}          & \multicolumn{1}{r|}{{\color[HTML]{009901} \textbf{51,646}}}                                 & \multicolumn{1}{r|}{227,634}                                 & \multicolumn{1}{r|}{{\color[HTML]{009901} \textbf{5,200}}} & {\color[HTML]{009901} \textbf{11,760}} & \multicolumn{1}{r|}{{\color[HTML]{009901}\textbf{62,672}}}                        & \multicolumn{1}{r|}{{\color[HTML]{009901}\textbf{5,273}}}                        & \multicolumn{1}{r|}{{\color[HTML]{009901} \textbf{57,519}}}                        & \multicolumn{1}{r|}{{\color[HTML]{009901}\textbf{7,336}}}                        & \multicolumn{1}{r|}{{\color[HTML]{009901}\textbf{62,981}}}                        &       {{\color[HTML]{009901}\textbf{24,864}}}                              \\ \hline
\end{tabular}
\end{adjustbox}
\end{table*}

\subsection{Main Results}
\label{eval:main-results}
In this subsection, we assess the performance of CAT using six benchmarks, considering a variety of tasks from regression to multi-class classification, and different types of data such as tabular data and images. We compare our method with the baselines above. Table~\ref{tab:main-results} shows the comparative results of different black-box and interpretable models averaged over three random seeds. Overall, both of the proposed CAT with order 2 and 3 comfortably outperform most of their interpretable counterparts and are comparable to some black-box DNNs on all six benchmarks. Since CAT and SPAM both utilize polynomial functions to make predictions, their performance is very close to other each. However, SPAM requires more model parameters in Tab.~\ref{tab:params}, since it leverages low-level features to make predictions while our CAT uses high-level concepts.

Below, we further examine the prediction performance of CAT models on each benchmark dataset. Firstly, our proposed method outperforms all interpretable baselines for the regression task on Airbnb dataset. The reason why CAT models works well for regression is that they aproximate real-valued target variables using Taylor polynomials, which can capture high-order interactions between different inputs. Particularly, CAT models, and SPAM which also use polynomials, have significantly lower prediction error (RMSE) than the other interpretable baselines that only utilize linear models. Also, CAT can compete closely with fully-connected MLPs, whose non-interpretable architecture implicitly models every order of feature interaction. Next, regarding binary classification of diabetes and recidivism, one notable observation is that using second-order concept interactions is sufficient for CAT to outperform all baselines including black-box models. One possible explanation is that our method can aggregates important patterns into its high-level concept representations, while other models can be negatively impacted by less useful low-level features. Additionally, on the multi-class classification dataset UCI-HAR, both CAT models outperform all interpretable baselines except for EBM. Although EBM performs better than CAT models and all other baselines on UCI-HAR, it is the least scalable interpretable method for this multi-class classification problem which we will further explore in the following experiment. Lastly, the proposed CAT is superior to interpretable baseline methods on both image datasets MNIST and CelebA, demonstrating its efficacy for visual reasoning.  

Furthermore, we benchmark the number of parameters and training throughput for all interpretable models on six datasets, which is summarized in Tab.~\ref{tab:params}. Here, we report the training throughput (measured in samples iterated per second) for methods that were trained on GPU machines over a fixed number of training iterations. Compared to other DNN-based interpretable methods, our CAT models generally have the smallest number of parameters and the highest training throughput. The main reason behind it is our adoption of high-level concept encoders to compress the input features, and the lightweight TaylorNet with Tucker Decomposition. Since NAMs use one DNN to encode each feature separately, they require a large number of parameters and take significantly more time to train. SPAMs also suffer from this problem as they adopt NAM to learn feature representations. Although Grand-Slamin utilizes sparse pairwise interactions to reduce the number of parameters compared to SPAMs, its training speed is not improved on large datasets such as Airbnb and Diabetes due to the inefficiency of soft decision trees~\cite{irsoy2012soft}. In addition, despite requiring more parameters than NBM for the UCI-HAR dataset, CAT models can be trained much more efficiently because our DNN concept encoders can be computed in parallel~\cite{dubey2022scalable}, while NBM uses one large DNN to learn many basis representations from the input features~\cite{radenovic2022neural}.
Even if EBMs are the most compact models for Airbnb and COMPAS datasets, one problem with EBM is the drastic increase in the number of parameters when they are applied on data with higher output dimension, whereas other methods including ours only seem to be affected by the input dimension. Specifically, going from Airbnb to Diabetes dataset, as the number of input and output dimensions increase from 126 to 176 and 1 to 2 respectively (Tab.~\ref{tab:data}), the number of parameters in EBM grows more than 12 times. Also, despite outperforming CAT and all other baselines on UCI-HAR, EBM requires the largest model of over 12 million parameters. Therefore, it is extremely difficult to scale EBM to multi-class classification datasets with hundreds of features in real-world settings~\cite{dubey2022scalable}.

\subsection{A Case Study of Interpretability}
\label{eval:case-study}
\noindent In this subsection, we present a case study to interpret the listing price prediction on the Airbnb dataset. Firstly, we provide a summary of the grouping of closely related features on Airbnb in Tab.~\ref{tab:airbnb_concepts}. Upon observing this table, it becomes apparent that even in the absence of concept annotation, if the features in the dataset are appropriately named and described in the metadata, general users can easily group them into human-understandable high-level concepts. It's important to note that our concept encoder differs from previous methods in concept-based learning~\cite{koh2020concept, chen2020concept, zarlenga2022concept}, where concepts are required to have not only meaningful names, but also accurate ground-truth values, for example 0/1 for \textit{bad/good} $Location$. In contrast, our concept encoder does not rely on precise ground-truth values. These concepts can then be effectively utilized by a white-box model for making predictions. During interpretation, the predictions can be traced back to these high-level concepts, enabling users to explain the model's decision at an abstract level.

\begin{table}[!thb]
\caption{Summary of high-level concepts extracted using feature names and provided metadata on the Airbnb dataset. This table demonstrates that the manual grouping of features into concepts is straightforward on a general application dataset with meaningful feature names and metadata.}
\label{tab:airbnb_concepts}
    \centering
    \resizebox{.9\columnwidth}{!}{%
    \begin{tabular}{|p{1.9cm}|>{\raggedleft}p{1.3cm}|>{\raggedleft\arraybackslash}p{4.4cm}|}
        \hline
        \textbf{High-level Concept} & \textbf{Num. of Features}&\textbf{Low-level Features} \\ \hline
        Amenities ($z_1$)&100& essentials, wifi, kitchen, tv, heating, washer, etc.\\ \hline
        Host ($z_2$)& 7 & host tenure, host location, response time, response rate, etc.\\ \hline
        Listing ($z_3$)&3& minimum nights, maximum nights, instant bookable\\ \hline
        Reviews ($z_4$)&7& overall, accuracy , clealiness, communication, value\\ \hline
        Property ($z_5$) &4& property type, room type, guests accomodated, bedrooms  \\ 
        \hline
        Location ($z_6$)& 5&neighborhood, district, city, longitude, latitude \\ 
        \hline
    \end{tabular}%
    }
\end{table}

\begin{figure}[!thb]
\centering
\includegraphics[width=.48\textwidth]{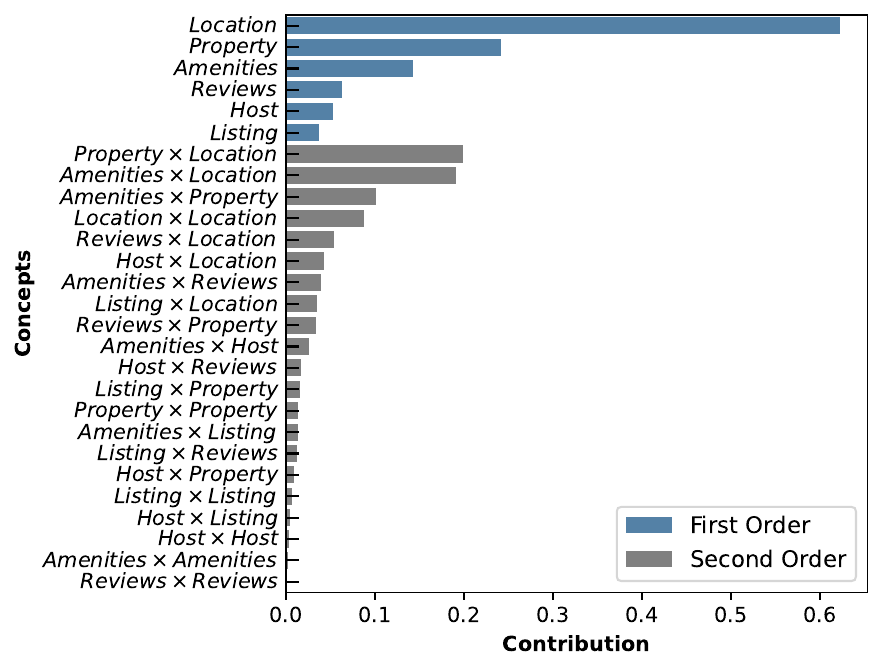}
\vspace{-1em}
\caption{Concept contributions using second-order CAT model for predicting listing price in the Airbnb dataset. Contributions are given by the standardized regression coefficients of the Taylor polynomial. We observe that the $Location$ and $Property$ descriptions influence the listing price the most.}
\label{fig:taylor_weights}
\vspace{-0.5em}
\end{figure}

The ease of interpretation for CAT models results from the combination of the compact high-level concept inputs and TaylorNet's inherrent interpretability. After learning the high-level concept representations by the concept encoders, we feed them into the white-box TaylorNet to model the non-linear functions with polynomials. For the Airbnb dataset, the discovered Taylor polynomial of order 2 that estimates the listing price is given by:

\begin{equation}\small
\begin{aligned}
    \vect{f}(\vect{z}) =& 0.02 z_{1}^{2} - 0.82 z_{1} z_{2} - 0.8 z_{1} z_{3} + 1.39 z_{1} z_{4} + 1.46 z_{1} z_{5} + 2.15 z_{1} z_{6} + 0.69 z_{1} \\&- 0.1 z_{2}^{2} + 0.48 z_{2} z_{3} - 1.01 z_{2} z_{4} - 0.33 z_{2} z_{5} - 1.08 z_{2} z_{6} - 0.46 z_{2} \\&+ 0.43 z_{3}^{2} - 1.0 z_{3} z_{4} - 0.75 z_{3} z_{5} - 1.2 z_{3} z_{6} - 0.44 z_{3} \\&+ 0.09 z_{4}^{2} + 0.96 z_{4} z_{5} + 1.07 z_{4} z_{6} + 0.45 z_{4} \\&+ 0.08 z_{5}^{2} + 2.28 z_{5} z_{6} + 0.96 z_{5} \\&- 0.46 z_{6}^{2} + 1.73 z_{6} - 0.03,
\end{aligned}
\label{eq:taylor_airbnb}
\end{equation}

\noindent where $z_m$ for $m=1,\dots,6$ is defined in Tab.~\ref{tab:airbnb_concepts}. The Taylor polynomial enables us to examine the contributions of concepts and their higher-order interactions using standardized regression coefficients~\cite{bring1994standardize}. These coefficients, akin to those in linear models, are computed by multiplying each polynomial coefficient by the standard deviation of its corresponding input feature and dividing by the standard deviation of the regression targets. Accordingly, we demonstrate the contributions of six high-level concepts and their second-order interactions to the listing price on the Airbnb dataset in Fig.~\ref{fig:taylor_weights}. From this figure, we observe that $Location$ and $Property$ description have the most significant influence on the price. Additionally, the second-order interactions between $Property \times Location$ and $Amenities \times Location$ are significant factors. This highlights that CAT's decision process mirrors human reasoning at a high level, as human individuals often attribute rental price to location, property quality, and amenities.

\begin{figure*}[!thb]
\centering
\includegraphics[width=\textwidth]{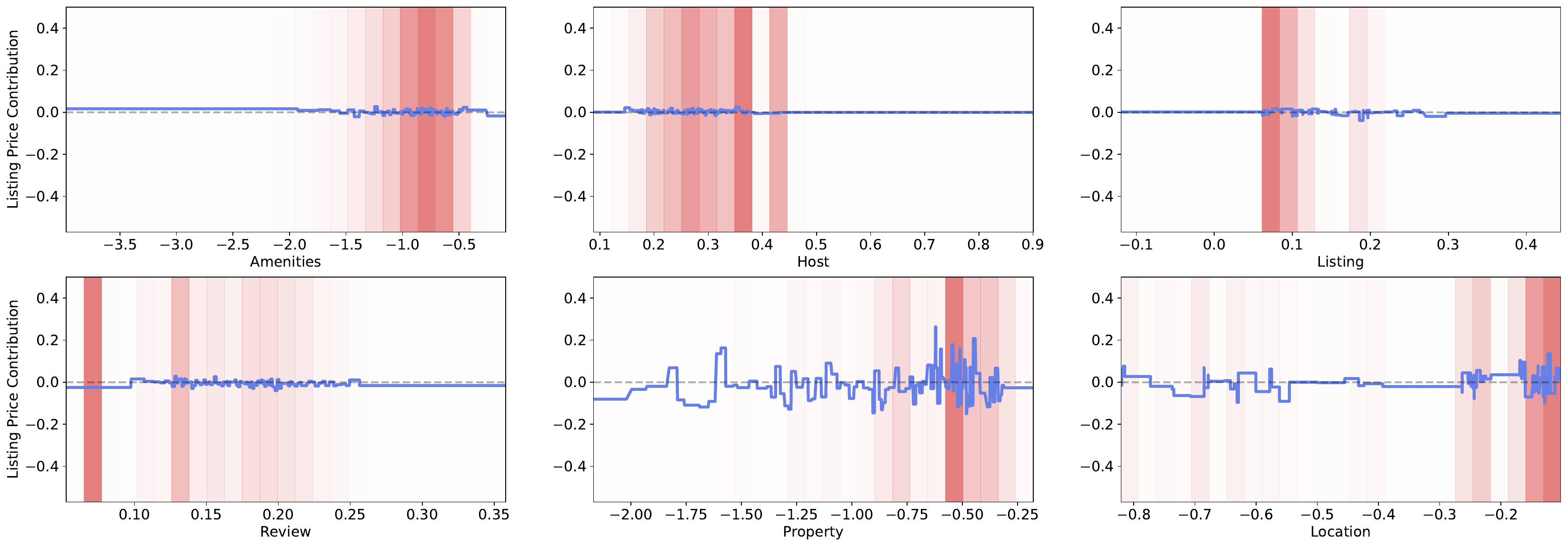}
\vspace{-2em}
\caption{Shape functions for the first-order concepts learned by the second-order CAT model on the Airbnb dataset. The x-axis represents the values of the concepts, while the y-axis indicates the contributions of each value to the listing price. The blue line represents the shape function for a concept. Pink bars represent the normalized data density for 25 bins of concept values.}
\label{fig:airbnb_shape}
\vspace{-1em}
\end{figure*}

Moreover, since CAT constructs explanations from a small number of concepts, the explanations are notably shorter than those from methods with feature-based interpretability, particularly those with high-order feature interactions like SPAM and EBM. Specifically, our second-order CAT explains the Airbnb listing price with 27 concepts and interactions, succinctly visualized in Fig.~\ref{fig:taylor_weights}, whereas second-order SPAM would necessitate 8127 terms.

Last but not least, for each concept learned by the second-order CAT model on the Airbnb dataset, we visualize its shape function and the corresponding normalized data density on the same graph, as illustrated in Fig.~\ref{fig:airbnb_shape}. In particular, the shape functions, indicated by the semi-transparent blue lines, elucidate how values of certain concepts, such as $Location$ and $Property$, influence the listing price. For example, from Fig.~\ref{fig:airbnb_shape}, values of the $Location$ concept within the range $[-0.58,-0.59]$ have a positive impact on the listing price. Consequently, users can discern the original features related to these concepts, enabling more detailed explanations.

In addition to this case study, we provide visualizations to interpret the gender prediction results on the image dataset CelebA using the second-order CAT model in Appendix~\ref{app:celeba}.

\subsection{Effectiveness of Concept Encoders}
To further demonstrate the effectiveness of our concept encoders, we integrate this component into two representative interpretable models: NAM and SPAM. Specifically, we extend NAM to NAM+ by incorporating the concept encoders to replace separate neural networks and subsequently feeding the concept representations into a linear model. Additionally, we develop SPAM+ as an extension of SPAM, wherein we substitute NAM with the concept encoders and then utilize the learned concept representations in a polynomial neural network. As illustrated in Table~\ref{tab:abl-poly}, both NAM+ and SPAM+ demonstrate superior performance when compared to NAM and SPAM, thereby emphasizing the effectiveness of our concept encoders. We also explore the impact of the concept encoders on the prediction performance and computational cost of the proposed CAT model in Appendix~\ref{app:encoder}.

\begin{table}[!thb]
\caption{Effectiveness of the concept encoders. Here boldface denotes where NAM+ and SPAM+ are better than their counterparts with feature-wise DNNs. The results averaged over three random seeds illustrate that incorporating our concept encoders improves the performance of the original models.}
\label{tab:abl-poly}
\vspace{-0.5em}
\centering
    \resizebox{.85\columnwidth}{!}{%
\begin{tabular}{|l|r|r|r|}
\hline
\multicolumn{1}{|c|}{}                                  & \multicolumn{1}{c|}{\textbf{AirBnb}}   & \multicolumn{1}{c|}{\textbf{Diabetes}} & \multicolumn{1}{c|}{\textbf{UCI-HAR}} \\ \cline{2-4} 
\multicolumn{1}{|c|}{\multirow{-2}{*}{\textbf{Models}}} & \multicolumn{1}{c|}{RMSE↓}             & \multicolumn{1}{c|}{Acc↑}              & \multicolumn{1}{c|}{Acc↑}                \\ \hline
NAM                                            & 0.6681                                 & 0.8242                                 & 0.9785                                   \\ \hline
NAM+                                            & \textbf{0.6069}                                & \textbf{0.8250 }                                & \textbf{0.9795}                                  \\ \hline
SPAM (Order 2)                                             & 0.5664                                 & 0.8230                                 & 0.9809                                   \\ \hline
SPAM+ (Order 2)                                    & \textbf{0.5515}                                & \textbf{0.8264}                                 & \textbf{0.9812}                                   \\ \hline
SPAM (Order 3)                                             & 0.5560                                 & 0.8272                                 & 0.9801                                 \\ \hline
SPAM+ (Order 3)                                    & \textbf{0.5504}                                 & \textbf{0.8274}                           & \textbf{0.9824}                                   \\ \hline
\end{tabular}%
}
\end{table}

\subsection{Hyper-parameters Tuning} \label{subsect:ablation}
We also investigate some important hyperparameters in TaylorNet, such as polynomial orders and ranks, on prediction performance, as detailed in Appendix~\ref{app:parameter}.
\section{Conclusion}
In this paper, we introduced CAT, a novel interpretable machine learning model capable of explaining and understanding predictions through human-understandable concepts. Unlike prior work, our method did not heavily rely on a large number of labeled concepts by domain experts. Specifically, the proposed CAT consisted of concept encoders that learned high-level concept representations from each group of input features, and a white-box TaylorNet that could approximate the non-linear mapping function between the input and output with polynomials. Additionally, Tucker decomposition was employed to reduce the computational complexity of TaylorNet. Extensive experimental results demonstrated that CAT not only achieved competitive or outstanding performance but also significantly reduced the number of model parameters. Importantly, it was able to explain model predictions using high-level concepts that humans can easily understand.

\clearpage


\bibliographystyle{ACM-Reference-Format}
\balance
\bibliography{ref}

\newpage
\appendix

 \onecolumn
\section{Appendix}
\subsection{Additional Interpretability Results on the CelebA Dataset}
\label{app:celeba}
\begin{figure}[!thb]
\centering
\includegraphics[width=0.70\textwidth]{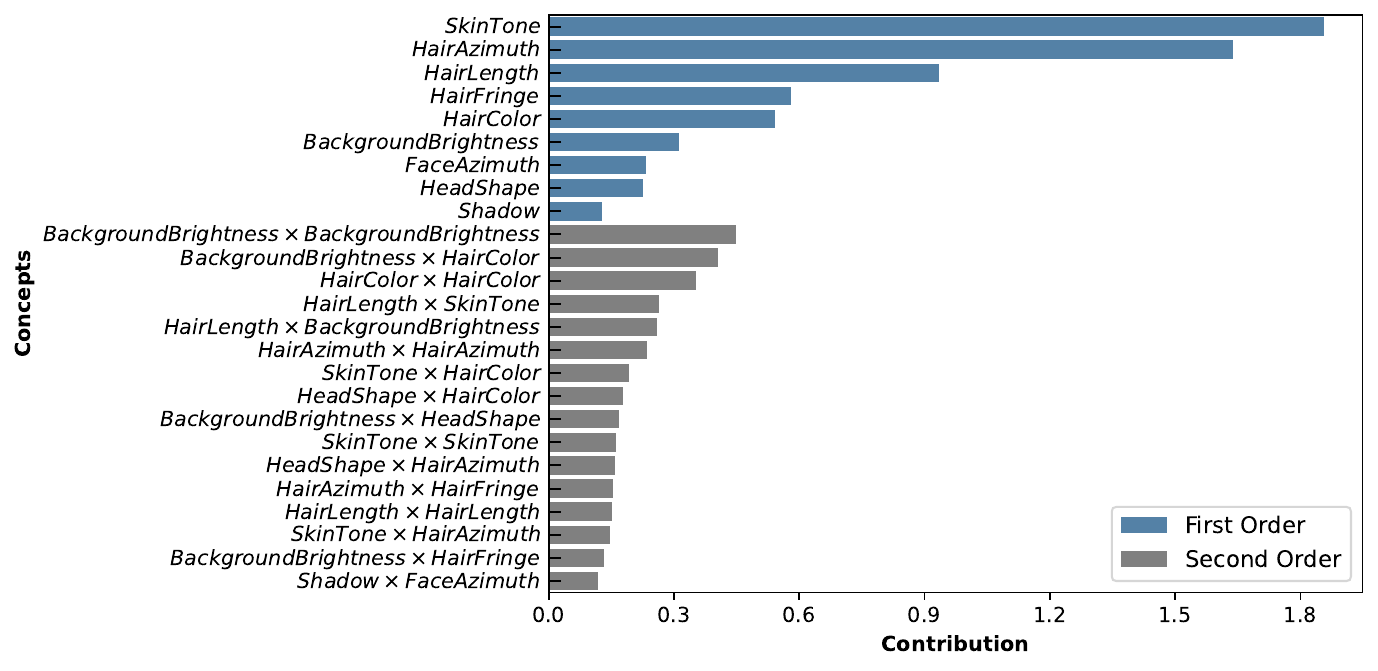}
\vspace{-1.5em}
\caption{Concept contributions using second-order Taylor for predicting gender in the CelebA dataset. Contributions are given by the standardized regression coefficients of the Taylor polynomial. We observe that the $Skin Tone$, $Hair Azimuth$, and $Hair Length$ concepts influence the gender prediction the most.}
\label{fig:celeba_weight}
\vspace{-1.8em}
\end{figure}

\begin{figure}[!thb]
\centering
\includegraphics[width=0.94\textwidth]{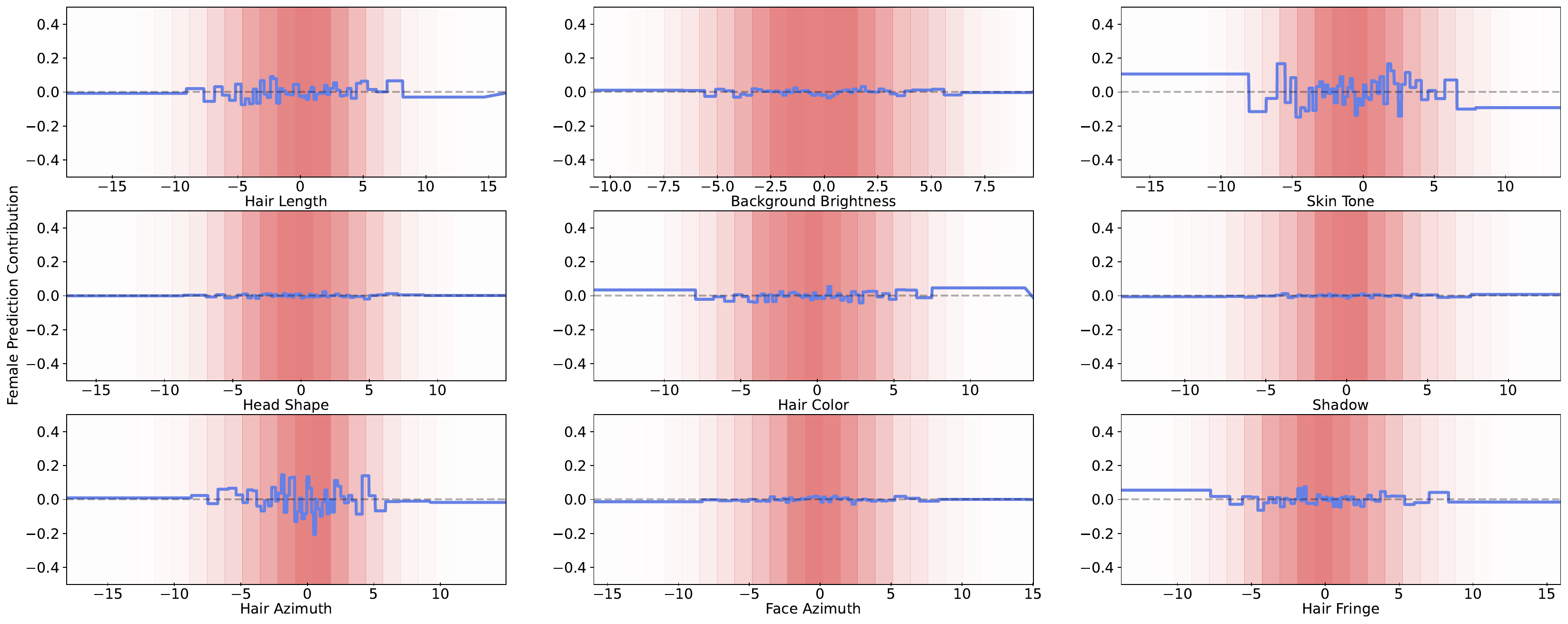}
\vspace{-1em}
\caption{Shape functions for the first-order concepts learned by the second-order CAT model on the CelebA dataset. The x-axis represents the values of the concepts, while the y-axis indicates the contributions of each value to the prediction of a female person. The blue line represents the shape function for a concept. Pink bars represent the normalized data density for 25 bins of concept values.}
\label{fig:celeba_shape}
\vspace{-0.5em}
\end{figure}

Besides the case study presented in Subsection~\ref{eval:case-study}, we provide additional visualizations to interpret the gender prediction results from the second-order CAT model on the image dataset CelebA. First, we showcase the gender prediction contributions of 9 high-level concepts acquired from disentangled representation learning, and the 16 most salient second-order interactions among these concepts in Fig.~\ref{fig:celeba_weight}. From this figure, it is evident that certain concepts such as $Skin Tone$, $Hair Azimuth$, and $Hair Length$ exhibit the most influence on the model's gender predictions, aligning closely with human reasoning at a high level. Furthermore, we include Fig.~\ref{fig:celeba_shape} to delineate the shape functions of 9 first-order concepts, providing clarity on how variations in specific concept values, such as $Skin Tone$ and $Hair Length$, impact predictions of female gender. For instance, individuals with darker $Skin Tone$ are more likely to be classified as male, whereas those with longer $Hair Length$ tend to be classified as female. Consequently, by employing the outlined concept-based explanation method, users can interpret the model's decision-making process with relative ease.
\vspace{-1em}

\subsection{Ablation of Concept Encoders in the Proposed CAT}
\label{app:encoder}
\begin{table*}[!thb]
\caption{Effectiveness of the concept encoders in the proposed CAT using three benchmark datasets. Here boldface denotes where the CAT models are better than their counterparts without concept encoders (CE). The results averaged over three random seeds demonstrate that the accuracy of CAT drops while the number of parameters increases without concept encoding.}
\label{tab:abl-encoder}
\vspace{-0.5em}
\centering
    \resizebox{0.75\textwidth}{!}{%
\begin{tabular}{|l|rr|rr|rr|}
\hline
\multicolumn{1}{|c|}{\multirow{2}{*}{\textbf{Models}}} & \multicolumn{2}{c|}{\textbf{Airbnb}}                                            & \multicolumn{2}{c|}{\textbf{Diabetes}}                                          & \multicolumn{2}{c|}{\textbf{UCI-HAR}}                                            \\ \cline{2-7} 
\multicolumn{1}{|c|}{}                        & \multicolumn{1}{c|}{RMSE↓}           & \multicolumn{1}{c|}{Num. of Params↓} & \multicolumn{1}{c|}{Acc↑}            & \multicolumn{1}{c|}{Num. of Params↓} & \multicolumn{1}{c|}{Acc↑}            & \multicolumn{1}{c|}{Num. of Params ↓} \\ \hline
CAT (Order 2)                                 & \multicolumn{1}{r|}{\textbf{0.5486}} & \textbf{48,742}                 & \multicolumn{1}{r|}{\textbf{0.8286}} & \textbf{51,310}                 & \multicolumn{1}{r|}{\textbf{0.9814}} & \textbf{161,138}                 \\ \hline
CAT w/o CE (Order 2)                          & \multicolumn{1}{r|}{0.5981}          & 131,136                         & \multicolumn{1}{r|}{0.8241}          & 138,288                         & \multicolumn{1}{r|}{0.9723}          & 194,256                          \\ \hline
CAT (Order 3)                                 & \multicolumn{1}{r|}{\textbf{0.5461}} & \textbf{52,990}                 & \multicolumn{1}{r|}{\textbf{0.8295}} & \textbf{51,646}                 & \multicolumn{1}{r|}{\textbf{0.9829}} & \textbf{227,634}                 \\ \hline
CAT w/o CE (Order 3)                          & \multicolumn{1}{r|}{0.5706}          & 860,670                         & \multicolumn{1}{r|}{0.8114}          & 869,580                         & \multicolumn{1}{r|}{0.9728}          & 1,190,656                        \\ \hline
\end{tabular}%
}
\end{table*}

\noindent To further assess the efficacy of the concept encoders (CE), we conduct an ablation study by systematically excluding them from the proposed CAT models (i.e., the input features are fed directly into TaylorNet), and measure the resulted prediction performance and computation cost using three datasets: Airbnb, Diabetes, and UCI-HAR. The results are presented in Table~\ref{tab:abl-encoder}. From this table, we can draw a conclusion that without the concept encoders, the accuracy of CAT drops while the number of parameters increases drastically. This underscores the necessity of incorporating them into our model.

\subsection{Hyper-parameter Tuning}\label{app:parameter}
\begin{figure*}[!thb]
\begin{center}
\begin{subfigure}[b]{0.30\textwidth}
    \centering
    \includegraphics[width=\textwidth]{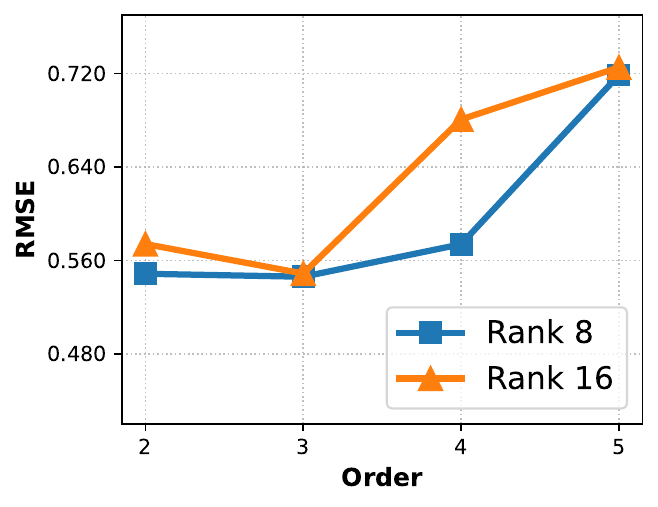}
    \vspace{-0.25in}
    \caption{Airbnb.}
    \label{fig:order_airbnb}
\end{subfigure}
\begin{subfigure}[b]{0.30\textwidth}
    \centering
    \includegraphics[width=\textwidth]{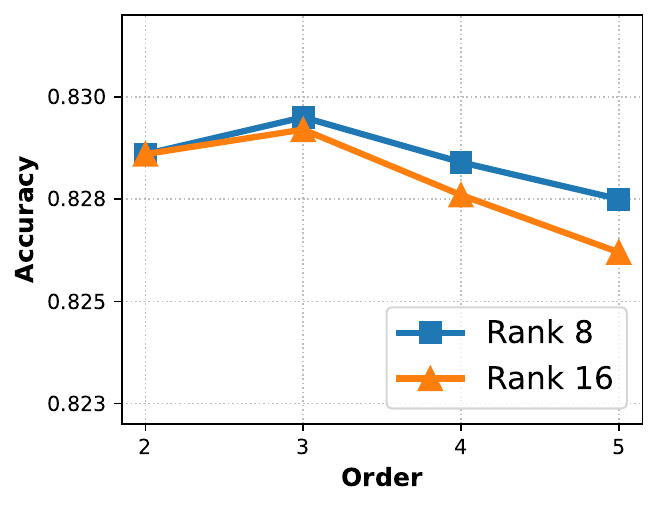}
    \vspace{-0.25in}
    \caption{Diabetes.}
    \label{fig:order_diabetes}
\end{subfigure}
\begin{subfigure}[b]{0.305\textwidth}
    \centering
    \includegraphics[width=\textwidth]{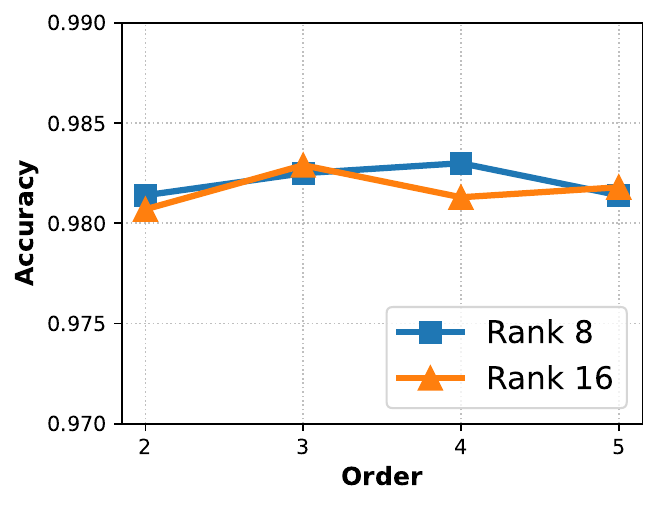}
    \vspace{-0.25in}
    \caption{UCI-HAR.}
    \label{fig:order_UCI-HAR}
\end{subfigure}
\end{center}
\vspace{-1em}
\caption{Effect of the order of TaylorNet on predictions using three benchmark datasets. Regarding the Airbnb dataset, lower RMSE scores indicate better performance. For Diabetes and UCI-HAR, we compare the prediction accuracy of the target categories. We can see that as the order of polynomials increases, the prediction performance on small datasets will drop due to overfitting.}
\label{fig:order_ablation}
\vspace{-2em}
\end{figure*}

\noindent\textbf{Effect of polynomial order.} We explore the impact of order $N$ in TaylorNet on model performance. Although it is natural to expect the quality of approximation to improve as the order of polynomials increases, the input and output dimensions are important factors for choosing an appropriate combination of polynomial orders and Tucker decomposition ranks to obtain satisfactory prediction performance. We can observe from Fig.~\ref{fig:order_ablation} that using higher-order Taylor polynomials (e.g. $\geq 3$) is beneficial on a dataset with higher input and output dimensions like UCI-HAR (Fig.~\ref{fig:order_UCI-HAR}), where the input space is approximately 10 times larger than the other two datasets. 

\begin{figure*}[!thb]
\begin{center}
\begin{subfigure}[b]{0.30\textwidth}
    \centering
    \includegraphics[width=\textwidth]{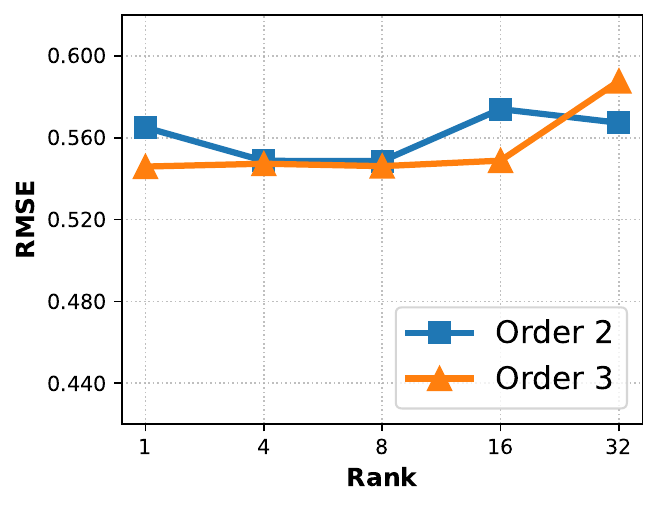}
    \vspace{-0.25in}
    \caption{Airbnb.}
    \label{fig:rank_airbnb}
\end{subfigure}
\begin{subfigure}[b]{0.30\textwidth}
    \centering
    \includegraphics[width=\textwidth]{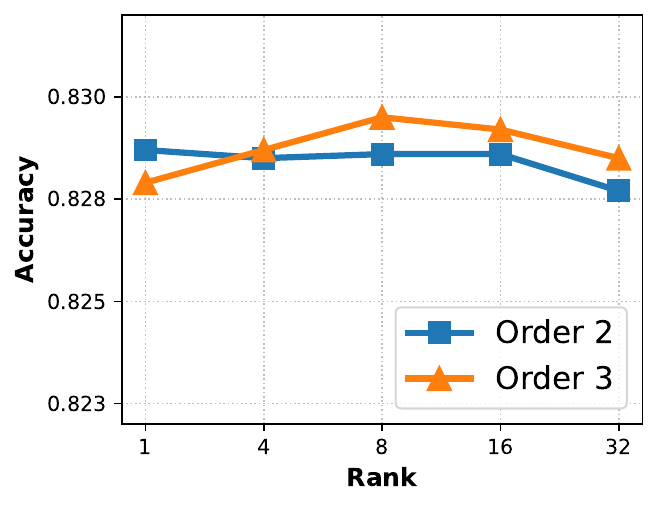}
    \vspace{-0.25in}
    \caption{Diabetes.}
    \label{fig:rank_diabetes}
\end{subfigure}
\begin{subfigure}[b]{0.305\textwidth}
    \centering
    \includegraphics[width=\textwidth]{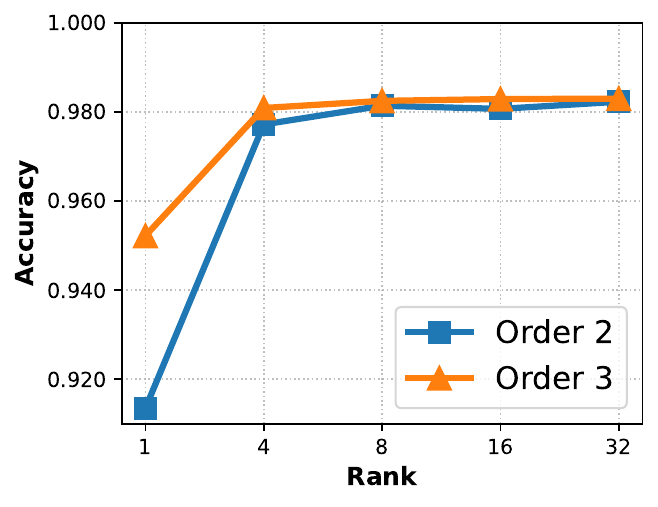}
    \vspace{-0.25in}
    \caption{UCI-HAR.}
    \label{fig:rank_UCI-HAR}
\end{subfigure}
\end{center}
\vspace{-0.15in}
\caption{Effect of the rank of Tucker decomposition on three benchmark datasets. Regarding the Airbnb dataset, lower RMSE scores indicate better performance. For Diabetes and UCI-HAR, we compare the prediction accuracy of the target categories. We can see that increasing the rank improves performance on UCI-HAR with large input dimension, but it will negatively impact performance on small datasets due to overfitting.}
\label{fig:rank_ablation}
\vspace{-1.5em}
\end{figure*}

\noindent\textbf{Effect of decomposition rank.} We also study the effect of the rank of Tucker decomposition on prediction performance. 
Additionally, choosing a higher rank $r$ for Tucker decomposition on smaller weight tensors on Airbnb and Diabetes datasets leads to diminishing returns. As illustrated in Fig.~\ref{fig:rank_ablation}, we can see the model performance will be gradually improved as the rank $r$ increases from $1$ to $8$, and then it will remain almost unchanged or degraded due to overfitting as the rank rises from 8 to 32.
\vspace{-1em}

\end{document}